\documentclass[letterpaper, 10 pt, conference]{ieeeconf}  

\IEEEoverridecommandlockouts                              

\usepackage{url}
\usepackage{amsmath}  
\usepackage{amssymb}  
\usepackage{mathtools} 
\let\labelindent\relax
\usepackage{enumitem}
\usepackage{graphicx}
\usepackage[table]{xcolor} 
\usepackage{booktabs}      
\usepackage{caption}       
\usepackage[compress]{cite}

\title{\LARGE \bf
TRACE: A Self-Improving Framework for Robot Behavior Forecasting with Vision-Language Models}

\author{Gokul Puthumanaillam$^{1*}$, Paulo Padrao$^{2*}$, Jose Fuentes$^{2*}$, Pranay Thangeda$^{1}$, \\ 
        William E. Schafer$^{1}$, Jae Hyuk Song$^{1}$, Karan Jagdale$^{3}$, Leonardo Bobadilla$^{2}$ and Melkior Ornik$^{1}$
\thanks{*Equal contribution}
\thanks{$^{1}$Gokul Puthumanaillam, Pranay Thangeda, William E. Schafer, Jae Hyuk Song, and Melkior Ornik are with the University of Illinois Urbana-Champaign (UIUC), Urbana, IL 61801, USA.
        {\tt\small \{gokulp2, pranayt2, wes6, jhsong2, mornik\}@illinois.edu}}%
\thanks{$^{2}$Paulo Padrao, Jose Fuentes, and Leonardo Bobadilla are with the Florida International University (FIU), Miami, FL 33199, USA.
        {\tt\small \{plope113, jfuen099\}@fiu.edu, bobadilla@cs.fiu.edu}}%
        \thanks{$^{3}$Karan Jagdale is with Lucid Group, Inc., Newark, CA, 94560.
        {\tt\small karanjagdale1@gmail.com}}%
        }

\begin{document}

\maketitle
\thispagestyle{empty}
\pagestyle{empty}

\begin{abstract}
Predicting the near-term behavior of a reactive agent is crucial in many robotic scenarios, yet remains challenging when observations of that agent are sparse or intermittent. Vision-Language Models (VLMs) offer a promising avenue by integrating textual domain knowledge with visual cues, but their one-shot predictions often miss important edge cases and unusual maneuvers. Our key insight is that iterative, counterfactual exploration--where a dedicated module probes each proposed behavior hypothesis, explicitly represented as a plausible trajectory, for overlooked possibilities--can significantly enhance VLM-based behavioral forecasting.
We present TRACE (\underline{T}ree-of-thought \underline{R}easoning \underline{A}nd \underline{C}ounterfactual \underline{E}xploration), an inference framework that couples tree-of-thought generation with domain-aware feedback to refine behavior hypotheses over multiple rounds. Concretely, a VLM first proposes candidate trajectories for the agent; a counterfactual critic then suggests edge-case variations consistent with partial observations, prompting the VLM to expand or adjust its hypotheses in the next iteration. This creates a self-improving cycle where the VLM progressively internalizes edge cases from previous rounds, systematically uncovering not only typical behaviors but also rare or borderline maneuvers, ultimately yielding more robust trajectory predictions from minimal sensor data.
We validate TRACE on both ground-vehicle simulations and real-world marine autonomous surface vehicles. Experimental results show that our method consistently outperforms standard VLM-driven and purely model-based baselines, capturing a broader range of feasible agent behaviors despite sparse sensing. 
Evaluation videos and code are available at \textcolor{teal}{\url{trace-robotics.github.io}}.        \end{abstract}

\section{Introduction}
Predicting the near-future behavior of a co-located agent is crucial in robotics \cite{horikawa2017previewed}, enabling an autonomous system to plan, adapt, or intervene in a timely manner. However, this goal is complicated by the realities of sparse observations: sensors often capture only partial or intermittent data \cite{beevers2006slam}, whether due to occlusions \cite{zhu2020occlusion}, stealth-related constraints \cite{puthumanaillam2024enhancing}, or simple bandwidth limitations \cite{puthumanaillam2024comtraq}. In such settings, contextual cues such as domain rules \cite{mehdipour2023formal} and environmental constraints \cite{thangeda2022expedited} become even more important for inferring an accurate portrayal of the agent's potential future actions.

Traditional behavior hypothesis generation methods typically rely on large, task-specific datasets \cite{brohan2022rt} or carefully hand-tuned dynamics \cite{leboutet2021inertial} and goal models \cite{dragan2013legibility}. While effective in well-explored scenarios, these methods can fail when observations are minimal or when the agent's maneuvers deviate from the training distribution--yet still abide by the underlying rules. Recently, vision-language models (VLMs) \cite{brohan2022rt} \cite{ahn2022can} have shown promise in mapping textual knowledge (e.g., domain guidelines, object attributes) onto visual input. 
VLMs offer promising potential for behavior hypothesis generation, explicitly represented as trajectories, due to their strong contextual understanding and reasoning capabilities. By integrating visual scene information with textual domain knowledge, they can interpret partial observations in light of known constraints. However, typical VLM deployments employ a single-step inference approach \cite{xu2024llava}, generating behavior hypotheses in isolation. Our key observation is that even when prompted to generate multiple hypotheses, VLMs tend to converge toward conservative or typical outcomes \cite{agrawal2018don}\cite{thrush2022winoground}, focusing on common or expected behaviors while overlooking physically plausible edge cases that may be crucial for planning and navigation.
\begin{figure}[t]
    \centering
    \includegraphics[width=0.45\textwidth, trim=22 40 20 35, clip]{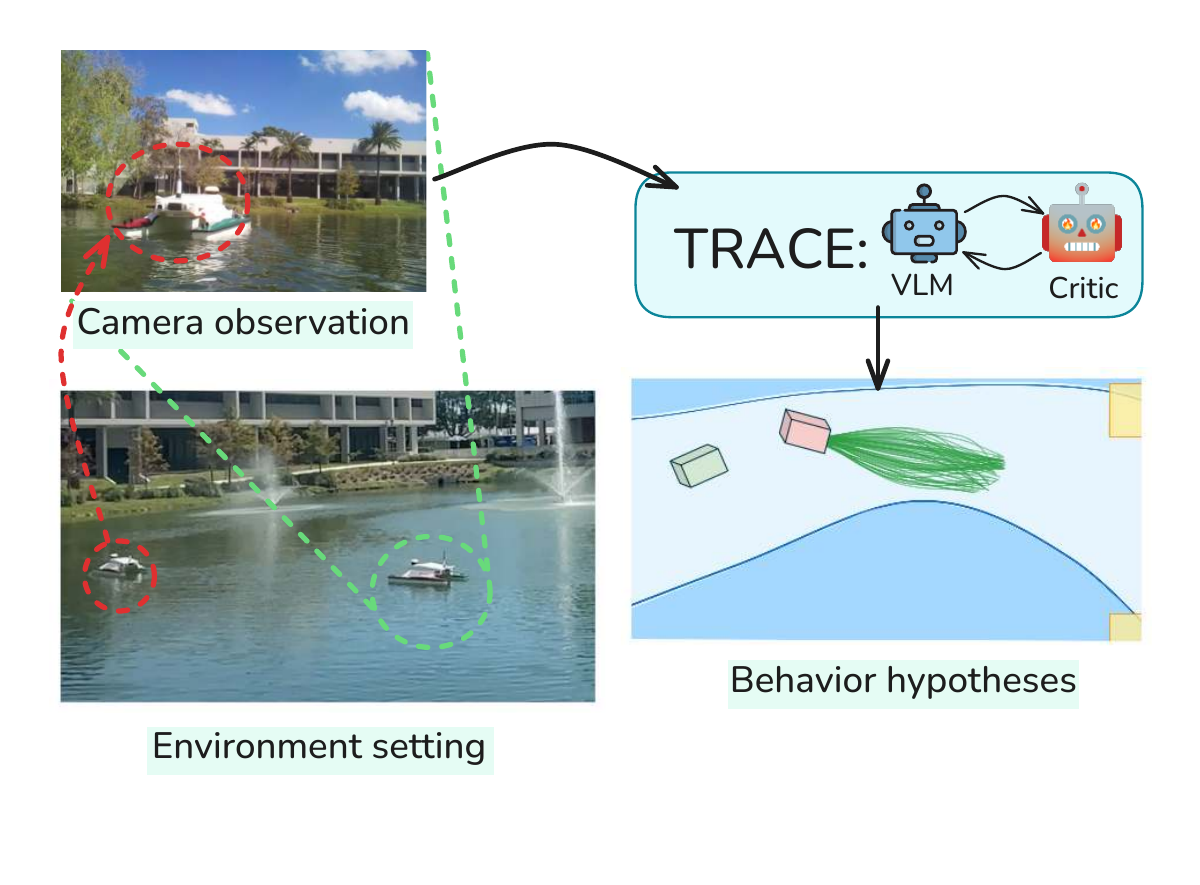}
    \caption{Illustration of TRACE applied to a marine navigation task. Top Left: The observer ASV sporadic observation. Top Right: Behavior hypotheses generated by TRACE.}
    \label{fig:arch-nav}
\end{figure}
The primary challenge with using VLMs for behavior hypothesis generation is thus twofold: ($i$) coping with sparse, partial observations that make the direct behavior hypothesis forecast uncertain, and ($ii$) extending VLM-based predictions beyond conservative or typical outcomes to account for uncommon yet feasible maneuvers. Our work addresses the question: Can systematically stress-testing VLM outputs through \emph{counterfactual} ``what-if" variants uncover low-probability yet feasible behavior hypotheses that would otherwise be overlooked?

Building upon this question, our paper presents \textbf{TRACE} (\textbf{T}ree-of-thought \textbf{R}easoning \textbf{A}nd \textbf{C}ounterfactual \textbf{E}xploration), an \emph{iterative inference framework} that employs a VLM, a world model, and a counterfactual exploration module in tandem. Rather than committing to a single round of predictions, TRACE expands a \emph{tree-of-thought}--generating candidate future trajectories across multiple steps, while a counterfactual critic highlights neglected edge cases still consistent with the partial observations. The accompanying world model enforces domain-specific constraints, filtering out nonsensical branches. By reintegrating these refined or newly discovered trajectories into subsequent iterations, TRACE creates a self-improving cycle where the VLM progressively internalizes edge cases from previous rounds, systematically building more comprehensive behavior hypotheses with each iteration. TRACE avoids the pitfalls of conservative or short-sighted forecasting and achieves broader, more robust trajectory coverage.

\textit{Statement of Contributions:}  
($i$) We introduce TRACE, an iterative tree-of-thought framework which fuses tree-of-thought reasoning with a counterfactual exploration component, 
($ii$) we propose a counterfactual critic with a world model that probes each candidate hypothesis, identifying plausible alternatives consistent with sparse sensing. We show how the feedback from the critic enables self-improvement in the VLMs, enabling it to progressively refine its predictions, learning from previous iterations to anticipate both common and rare behavioral patterns better,
($iii$) we highlight the platform-agnostic nature of TRACE by demonstrating its utility across diverse robotic navigation settings in both hardware and simulation, showing that even with partial sensor feedback, the proposed approach significantly the breadth of behavior hypotheses compared to conventional VLM-based or classical model-based baselines.

While our work relates to reachability analysis \cite{shafa2022reachability}\cite{meng2024online}, we avoid this formal approach due to its requirement for precise dynamic models, difficulty handling complex rules, and computational limitations. Instead, we specifically focus on VLMs within an iterative framework that leverages counterfactual exploration to demonstrate how VLMs can progressively learn from previous iterations, generating comprehensive behavior hypotheses that cover both common and edge-case navigation maneuvers.


\section{Related Work}
\paragraph{Behavior Hypothesis Generation in Robotics}
Behavior hypothesis generation has evolved from classical motion models \cite{bennewitz2002learning} to data-driven approaches \cite{korbmacher2022review}. Early works utilized plan recognition with hidden variables \cite{baker2014modeling}, while recent methods leverage learning for trajectory forecasting \cite{puthumanaillam2024tab}. Generative adversarial networks have emerged as effective tools for modeling physically feasible trajectories \cite{sadeghian2019sophie}, incorporating diversity loss mechanisms to capture the multimodal nature of agent movement. Recurrent Neural Network-based approaches \cite{deo2018multi}, particularly those with social pooling mechanisms \cite{vemula2018social}, have shown success in encoding interpersonal dynamics. Graph neural networks further enhance relational reasoning by constructing spatio-temporal representations \cite{mohamed2020social} that capture contextual cues through graph convolutions.

\paragraph{Model-Based Approaches}
Model-based behavior hypothesis generation techniques posit explicit models of agent decision-making. Reinforcement learning (RL) approaches \cite{puthumanaillam2024comtraq} \cite{chung2025predicting} model agents that optimize reward functions, enabling future behavior prediction through policy simulation \cite{AlSharman2023}. When agent objectives are unknown, Inverse Reinforcement Learning (IRL) \cite{ng2000algorithms} infers reward functions from observed behaviors. 
Probabilistic graphical models offer another approach \cite{li2019dynamic}, treating agent behavior as a latent variable for Bayesian inference \cite{hernandez2024bayesian}. POMDP formulations \cite{qiao2021learning} continuously update belief distributions on possible goals while accommodating uncertainty. In multi-agent domains such as autonomous driving, hybrid approaches combine hand-crafted maneuver models with learned components \cite{Wang2025}, using hidden mode representations for driver intentions. These methods present distinct trade-offs in long-horizon forecasting capabilities, incentive modeling requirements, and complexity management with multiple hypotheses. With sparse observations, simple model-based predictors can outperform complex learned models but fail when behaviors deviate from simplistic assumptions \cite{lefevre2014survey}.

\paragraph{VLMs for Behavior Hypothesis Generation}
VLMs have recently been applied to behavior forecasting in robotics, bringing semantic knowledge to visual processing \cite{bao2024handsonvlm}\cite{zhao2024vlmpc}. Language-based trajectory prediction approaches transform the forecasting task into natural language problems \cite{ma2024vision}, encoding past coordinates and scene descriptions for language model completion. These methods have outperformed traditional regression models by taking advantage of contextual cues and common-sense knowledge acquired during pre-training \cite{bae2024can}. 
Despite their promise, VLM-based behavior hypothesis generators face several limitations \cite{nasiriany2024pivot}\cite{feng2025reflective}. Current implementations typically lack iterative refinement mechanisms, operating in an open-loop fashion that requires re-prompting and human-generated prompts for dynamic environments \cite{guruprasad2024benchmarking, tsao2024trajprompt}. VLMs also heavily depend on their training data distribution, potentially failing in scenarios underrepresented in pre-training. 
\label{sec:trace}

\begin{figure*}[httb]
    \centering
    \includegraphics[width=0.9\textwidth, trim=32 90 32 22, clip]{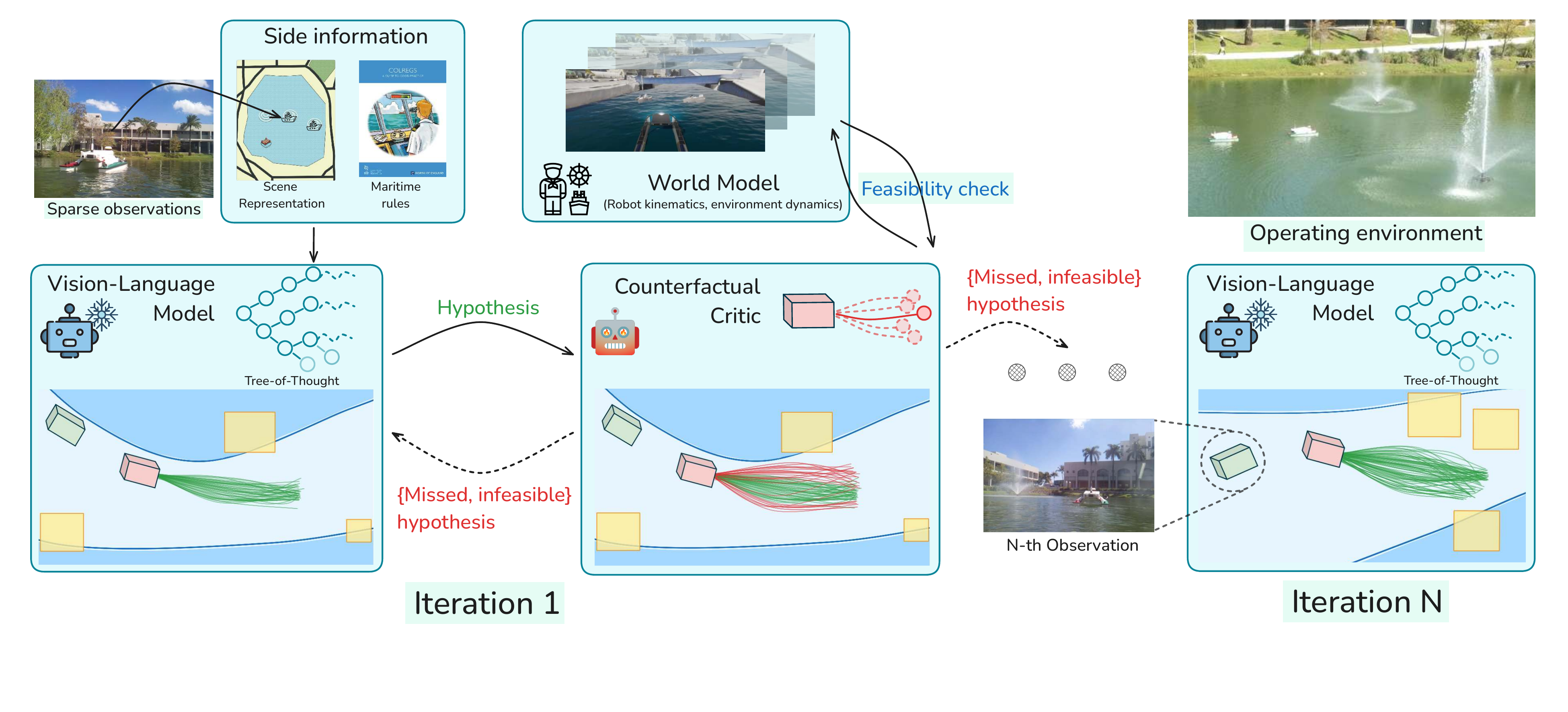}
    \caption{TRACE operates through an iterative three-component cycle: ($i$) Hypothesis Generation, where a VLM analyzes sparse observations to propose initial behavior hypotheses; ($ii$) Counterfactual Exploration, where a critic identifies overlooked edge-cases; ($iii$) Self-Improvement integrates both valid and rejected hypotheses into the VLM context for enhanced predictions.}
    \label{fig:arch}
\end{figure*}

In contrast to previous research, we place our research at the intersection of VLMs, reasoning, and counterfactual analysis for robotic behavior prediction. While existing approaches typically operate in isolation—either relying solely on model-based techniques or applying VLMs in a one-shot manner—our framework creates a self-improving cycle where the VLM progressively internalizes edge cases from previous rounds through counterfactual feedback, systematically expanding beyond the limitations of both traditional model-based methods and standard VLM implementations.

\section{Problem Formulation}
This section formalizes the behavior hypothesis generation problem under sparse observations and contextual constraints.
We focus on a navigation scenario with two agents: an \emph{observer} and a \emph{target} whose behavior we wish to predict.  We define a behavior hypothesis as a plausible trajectory of the target agent. Let time be discretized as \(t = 1, 2, \dots, T\) and \(s(t)\) be the full state of the target agent at time \(t\), which could include position, velocity, or other navigation-relevant features (e.g., steering angle). We \emph{do not} assume access to the target's internal control policy or full historical trajectory; instead, at certain times \(t\) where the observer obtains sensor data, we receive an \emph{intermittent} measurement \(\Omega(t)\). For instance, \(\Omega(t)\) could be a noisy position estimate or a camera-based target detection. In many real-world settings--especially those involving occlusions, sensor range limits, or stealth-based constraints--observations may be missing for several time steps, underscoring the difficulty of maintaining accurate, continuous state estimates. 
We assume the observer has a high-level static representation of the environment (e.g., satellite imagery) \(\mathrm{env}\) that underpins the operating environment. This assumption is reasonable for many real-world navigation tasks, where environmental layouts (e.g., roads) are readily available through mapping services.

Whenever a new observation \(\Omega(t)\) arrives, the observer's goal is to generate a set of plausible future trajectories:
\begin{align*}
\Gamma(t) 
\;=\;
\bigl\{
(s(\tau))_{\tau=t}^T :
\underbrace{\mathcal{F}(s(\tau),s(\tau+1))=1}_{\text{feasibility constraints}}, \underbrace{\text{matches } \Omega(t)}_{\text{ consistency}}
\bigr\},
\end{align*}
where each candidate in \(\Gamma(t)\) is a sequence \((s(t), s(t+1), \dots, s(T))\).
Let \(\mathcal{C}\) denote the constraints (e.g., restricted zones or rules of navigation). A \emph{world model} (described in Section~\ref{sec:world-model-implementation}) encodes these constraints and simulates valid transitions from one state to the next. A trajectory \((s(\tau))_{\tau=t}^T\in \Gamma(t)\) iff \(\mathcal{F}\bigl(s(\tau), s(\tau+1)\bigr) = 1\) and  \(s(\tau)\) does not contradict  \(\Omega(t) \forall \tau = t, \ldots, T\). 
Thus, the primary challenge is to maintain \(\Gamma(t)\) so that it:
($i$) \emph{Respects feasibility:} Each candidate trajectory is physically plausible within the world model.
($ii$) \emph{Maintains consistency:} No candidate breaks alignment with sensor observations or domain knowledge.
($iii$) \emph{Captures heterogeneity:} Instead of converging to a single “most likely” path, \(\Gamma(t)\) must include \emph{multiple} plausible trajectories, including rare but valid maneuvers.

\section{TRACE: Tree-of-Thought Reasoning and Counterfactual Exploration}
Having established the problem statement, we now present our approach. TRACE integrates a VLM within an iterative framework that explores and refines behavior hypotheses through a tree-of-thought enhanced by counterfactual critic. 
\subsection{Vision-Language Tree-of-Thought Inference}
\label{sec:vlm-tot}
The VLM serves as the core hypothesis generator, leveraging its ability to reason over visual inputs and domain knowledge. 
At any discrete time \(t\) when a new measurement \(\Omega(t)\) arrives, a transform is applied to obtain an approximate position \(\widehat{s}(t)\) of the target agent in the environment representation \(\mathrm{env}\). For instance, a camera snapshot can provide a bounding box of the agent, which is projected onto the top-down map via known calibration parameters \cite{zhang2002flexible}. Since the transformation depends on the sensor being used and many prior works have focused on this standard transformation, we do not explicitly detail it in our paper.
Since the observer's position is known, this transformation establishes a common spatial reference frame where both agents' states are represented within \(\mathrm{env}\). Fig. \ref{fig:arch} (side information) shows an example of the representation.
The updated environment \(\mathrm{env}\), now comprises a high-level map, the target's approximated state \(\widehat{s}(t)\) annotated on that map, along with the observer's position. The combination of visual data \(\mathrm{env}\), textual context (\(\mathcal{C}\)), and the latest sensor reading (\(\Omega(t)\)) serve as contextual side-information for the VLM's reasoning process.


\textit{Tree-of-Thought Generation:}
Given the annotated environment map \(\mathrm{env}\) and \(\widehat{s}(t)\), the VLM proposes a branching set of short-horizon next states
$ \bigl\{\widetilde{s}_1,\widetilde{s}_2,\dots,\widetilde{s}_k\bigr\}$
spanning times \(t+1, t+2, \dots, t+\Delta\). Each \(\widetilde{s}_{j}\) represents a possible future position, subject to basic feasibility constraints inferred from the available context. Specifically, we prompt the VLM to analyze the map layout to avoid obstacles and restricted zones while adhering to domain rules (\(\mathcal{C}\)). The VLM accomplishes this to the extent these constraints are encoded in its learned representation. Formally,
\[
\mathcal{G}_{\mathrm{VLM}}
\bigl(\mathrm{env}, \widehat{s}(t), \Omega(t)\bigr)
=
\bigl\{\widetilde{s}_{1}, \dots, \widetilde{s}_{k}\bigr\}.
\]
Subsequent expansions apply the same operator to each proposed state $\widetilde{s}_{j}
\mapsto (\widetilde{s}_{j,1},\dots,\widetilde{s}_{j,m})$.
Repeating this process to depth \(\Delta\) yields a branching tree of partial trajectories, thereby retaining multiple lines of progression rather than collapsing prematurely to a single highest-likelihood path.

\subsection{Counterfactual Critic}
\label{sec:counterfactual}
A single tree-of-thought expansion often concentrates on typical or high-probability outcomes \cite{agrawal2018don} inferred by the VLM. However, unusual or borderline behaviors may also satisfy the domain constraints and observations, and missing them can undermine the robustness of the predicted behaviors. We address this by generating \emph{alternative} behavior hypotheses near the VLM's outputs, then filtering them through the world model to ensure feasibility.

\textit{Critic overview: }Our critic network \(\mathcal{N}_\phi\) takes as input the trajectory proposed by the VLM  \(\Xi\), a local embedding of the environment map \(\mathrm{env}\), and relevant constraints \(\mathcal{C}\). It outputs a set of incremental state offsets, $\delta = \bigl(\delta_1,\,\ldots,\,\delta_\Delta\bigr)$,
where each \(\delta_k\) captures a targeted perturbation to \(\widetilde{s}_{t+k}\). Combining these perturbations produces a counterfactual trajectory,
$\Xi^* \;=\; (\widetilde{s}_{(t+1)} + \delta_1,\,\ldots,\,\widetilde{s}_{(t+\Delta)} + \delta_\Delta)$.

\paragraph{Forward Pass and Feasibility Filtering}
For each partial trajectory \(\Xi\), the module performs a forward pass through \(\mathcal{N}_\phi\) to obtain \(\Xi^*\). The proposed counterfactual is immediately evaluated by the world model (detailed in Section~\ref{sec:world-model-implementation}), which examines \(\Xi^*\) against the feasibility function \(\mathcal{F}\). Any violation--such as collision or kinematic constraint breach--disqualifies \(\Xi^*\). Trajectories that satisfy \(\mathcal{F}\) are incorporated into the tree structure, effectively expanding the hypothesis space beyond the VLM's initial predictions.

\paragraph{Input Representation}
The perturbation network processes two components:
($i$) \textit{Local Trajectory Context:} An embedded representation of \(\Xi\), encoding the VLM's proposed states \(\widetilde{s}(t+k)\).  
($ii$) \textit{Environment Embedding:} A spatial encoding from \(\mathrm{env}\) centered around \(\Xi\), combined with a latent representation of the textual constraints \(\mathcal{C}\).
These components are integrated through cross-attention within the network to generate contextually relevant offsets \(\delta\).

\paragraph{Training Criterion}
The network \(\mathcal{N}_\phi\) is trained to produce trajectories that are both feasible and substantively different from the VLM's baseline. We define the loss:
\[
\mathcal{L}(\Xi, \Xi^*)
\;=\;
\alpha \,\mathcal{L}_\text{feas}(\Xi^*)
\;+\;
\beta \,\mathcal{L}_\text{div}(\Xi, \Xi^*),
\]
where \(\mathcal{L}_\text{feas}(\Xi^*)\) penalizes infeasible transitions identified by the world model (\(\mathcal{F}=0\)) or contradictions with sensor data \(\Omega(t)\), while \(\mathcal{L}_\text{div}(\Xi, \Xi^*)\) rewards meaningful divergence from the baseline trajectory. The hyperparameters \(\alpha\) and \(\beta\) balance these competing objectives, ensuring that counterfactuals remain physically plausible while introducing novel maneuver patterns that might otherwise be overlooked.

\subsection{Integration into the Iterative Reasoning Loop}
\label{sec:integration-loop}
The counterfactual trajectories generated in Section~\ref{sec:counterfactual} are systematically integrated back into the reasoning process to create a self-improving cycle for the VLM. After world model verification, each trajectory node is classified into three categories: feasible hypotheses, physically implausible paths, and overlooked edge cases providing valuable learning signals for the VLM in subsequent iterations.

\begin{figure}[ht]
    \centering
    \includegraphics[width=0.49\textwidth , trim=10 35 10 31, clip]{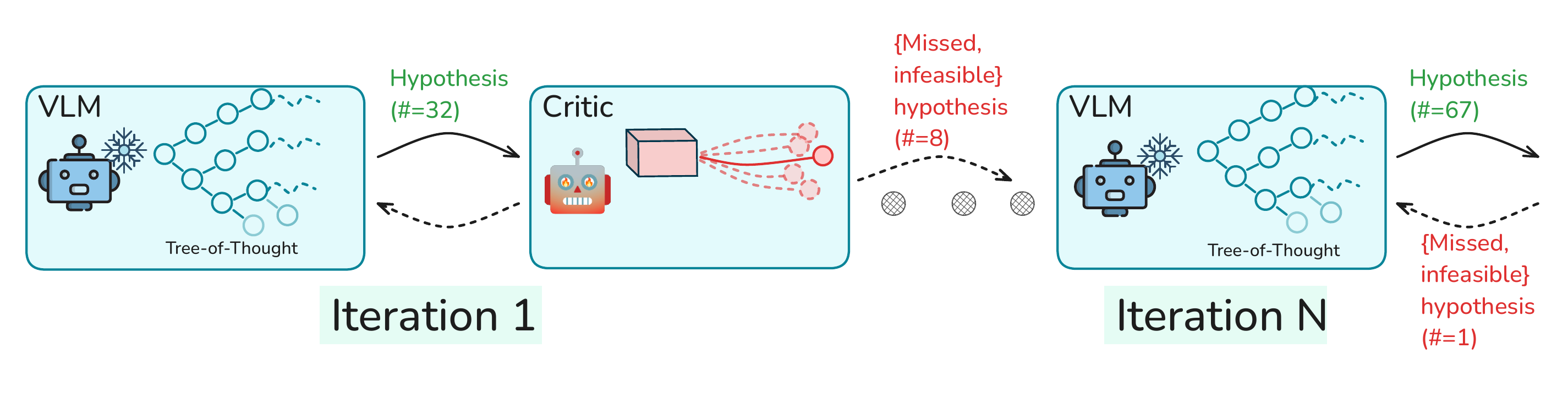}
    \caption{Abstraction of the self-improvement cycle.}
    \label{fig:wide-figure}
\end{figure}

This feedback mechanism creates a powerful learning dynamic: the VLM receives explicit examples of both successful hypotheses and failure modes from previous iterations. By including these examples in the context window for subsequent prompts, the model implicitly learns behaviors that are physically plausible and which edge cases it previously missed. Unlike traditional fine-tuning approaches, this learning happens during inference time without parameter updates. The VLM progressively internalizes domain constraints and common failure patterns, generating increasingly comprehensive behavior hypotheses with each iteration.
The self-improvement cycle continues until new sensor data \(\Omega(t')\) arrives, TRACE prunes inconsistent branches. Through each iteration, not only does the coverage of possible agent behaviors grow more comprehensive, but the VLM itself becomes more effective at generating valid hypotheses. Its increasing capabilities creates a compounding benefit where later iterations require fewer counterfactual corrections as the model learns to anticipate both common behaviors and previously identified edge cases. This flow between VLM reasoning and the critic effectively transforms the VLM from a generic predictor into a domain-aware behavior hypothesis generator specifically attuned to the physical and contextual boundaries of the current navigation scenario.

\subsection{World Model Implementation}
\label{sec:world-model-implementation}

The world model enforces physical, kinematic, and regulatory constraints, ensuring that each proposed state transition remains feasible. Though implementations vary by domain, our framework relies on two components:

\paragraph{Kinematic Feasibility Module}
This component verifies whether state transitions respect motion limits such as maximum velocity, turning radius, and acceleration bounds. Implemented as a lightweight physics function, it evaluates each candidate step against the agent's operational parameters and rejects physically impossible maneuvers.

\paragraph{Domain Compliance Validator}
Domain-specific rules and constraints \(\mathcal{C}\) are implemented as boundary checks that detect collisions with obstacles, violations of navigable zones, or breaches of right-of-way regulations. This module employs geometric intersection tests and rule-based verification to maintain compliance with environmental restrictions.

These components define a binary feasibility function $\mathcal{F}$ returning $1$ for valid transitions and $0$ otherwise. Candidate trajectories are retained only if all transitions satisfy $\mathcal{F}$, pruning infeasible branches and validating novel paths.

\section{Experiments and Results}
\label{sec:experiments}

\begin{figure*}[t]
    \centering
    \includegraphics[width=\textwidth]{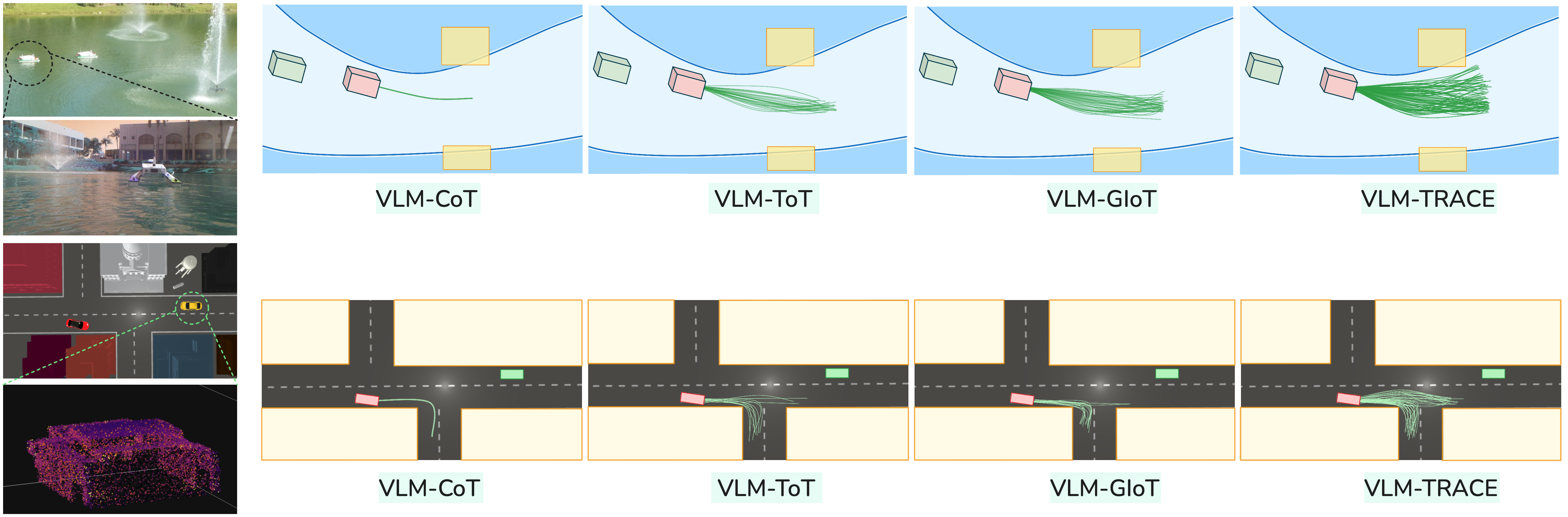}
    \caption{Comparison of behavior hypotheses generated by TRACE versus baselines for T2 and T5 in hardware experiments and simulation. Green indicates the observer, red shows the target, and green lines represent the generated hypotheses.}
    \label{fig:qual-figure}
\end{figure*}

\textit{Motivation:} We evaluate TRACE in scenarios involving two agents, an observer and a target, operating in a shared environment. Due to real-world constraints (such as limited sensor range or communication bandwidth restrictions), the observer receives only sporadic measurements of the target agent's state. Our goal is to predict the target's behavior hypotheses--trajectories--despite this measurement sparsity, enabling effective decision-making by the observer.

\subsection{Robot Setup}
\label{sec:robot-setup}

\paragraph{Hardware Experiments (Marine Navigation)}
Our real-world in-the-wild tests utilize two SeaRobotics Surveyor Autonomous Surface Vehicles (ASVs) operating in a lake containing multiple fountains that serve as both physical obstacles and sources of water disturbance. The environment contains designated safe navigable channels analogous to maritime lanes and restricted areas (fountains and shorelines) that must be avoided. Though less rigidly defined than road networks, these navigational constraints enforce a structured space within which both ASVs must operate.
The ASV's action space is defined by \(a = (\lambda, \alpha\)), where \(\lambda \in [0,\lambda_{\max}]\) represents propulsion torque (\(\lambda_{\max} = 85\) N·m), \(\alpha \in [0,2\pi)\) indicates steering angle. The maximum operational speed is 2 knots. The ASVs are equipped with GPS for localization. The observer ASV, receives camera snapshots of the target ASV at irregular intervals. Between camera observations, the target's state is entirely unknown.

\paragraph{Simulated Experiments (Autonomous Driving)}
To evaluate TRACE under controlled conditions, we developed a custom medium-fidelity simulator featuring a two-agent driving scenario on a grid of roads and buildings in an urban setting. The observer vehicle tracks a target vehicle using simulated LiDAR readings obtained at sparse intervals.
The vehicles operate with a maximum speed of 15 m/s, a turning radius of 5 meters, and maximum acceleration/deceleration of $\pm$3 m/s$^2$. Environmental constraints include road boundaries, building footprints, traffic rules, and intersection protocols, providing well-defined restrictions.

Our experiments use GPT-4 Turbo with Vision via the OpenAI API, though TRACE's architecture is model-agnostic and compatible with any vision-language model.

\subsection{Tasks}
\label{sec:tasks_baselines}
\subsubsection{Maritime Scenarios}
\label{sec:tasks}
We evaluate TRACE in distinct maritime navigation settings, each involving specific COLREG (International Regulations for Preventing Collisions at Sea) \cite{colreg2003convention} rules that govern the behavior of the agents:

\begin{itemize}[leftmargin=1.5em]
    \item T1 -- \emph{Head-On Encounter (Rule 14)}: The observer and target ASVs approach each other on approximately reciprocal courses. This task tests TRACE's ability to predict whether the target will alter course to starboard as prescribed by maritime regulations.
    
    \item T2 -- \emph{Overtaking in a Narrow Channel (Rule 13)}: The target ASV navigates within a confined waterway while the observer follows. This task evaluates prediction quality when the target has limited lateral maneuver space.
    
    \item T3 -- \emph{Crossing Situation (Rule 15)}: The observer and target approach an intersection of navigable channels at approximately right angles. This task tests the ability to predict whether the target, as the stand-on vessel, will maintain course or yield.
\end{itemize}

\subsubsection{Ground Vehicle Scenarios}
In simulation, we evaluate on tasks adapted to ground vehicle navigation:

\begin{itemize}[leftmargin=1.5em]
    \item T4 -- \emph{Overtaking on a Two-Lane Road}: The observer follows the target on a bidirectional road segment. This setting tests TRACE's ability to predict the target's behavior: whether it will employ longitudinal control, lateral control, or a combination of both.
    
    \item T5 -- \emph{Right-Turn Decision Point} (T5): The target approaches an intersection where it can either continue straight or execute a free right turn. This task evaluates the ability to predict directional intent from subtle cues such as vehicle positioning before the actual turn.
\end{itemize}

We measure TRACE and the baselines' ability to generate a comprehensive set of plausible behaviors of the target, despite having only intermittent observations.

\subsection{Baselines.}
We evaluate four main inference strategies, each granted access to the same world model for feasibility checks:

\noindent\textit{Chain-of-Thought (CoT) VLM \cite{cot}:} A sequential reasoning approach where the model articulates its step-by-step logic to predict a single trajectory for the target agent. CoT represents the standard prompted reasoning baseline, lacking the ability to generate multiple hypotheses.

\noindent\textit{Guided Iteration of Thought (GIoT) VLM\cite{giot}:} A controlled iterative process that refines predictions over a predefined number of steps. In GIoT, the model continues generating new responses for $N-1$ iterations without concluding early, only providing its final set of trajectory hypotheses in the $N$-th iteration. This approach ensures an extensive exploration of the solution space while maintaining multiple hypotheses.
    
    \noindent\textit{Tree-of-Thought (ToT) VLM\cite{tot}:} A branching model that produces a short-horizon tree of possibilities (Section~\ref{sec:vlm-tot}) through explicit exploration of multiple reasoning paths.

\subsection{Metrics}
Since the underlying environment is fully observable offline, we construct a ground-truth set of all physically and contextually feasible trajectories \(\Gamma^\star\) for each scenario using the world model. We then measure the \emph{coverage ratio}:
\[
\mathrm{Coverage}(\Gamma^\dagger) 
\;=\;
\frac{|\Gamma^\dagger \,\cap\, \Gamma^\star|}
     {|\Gamma^\star|},
\]
where \(\Gamma^\dagger\) denotes the set of trajectories output by a given approach. Higher coverage indicates that more of the truly valid possible future maneuvers are correctly identified.

\subsection{Key Results}
\label{sec:results}

\textit{Key finding 1: }\textbf{Counterfactual exploration expands the behavioral hypothesis space.}
We evaluate each method's ability to generate a comprehensive set of feasible trajectories against the ground-truth set \(\Gamma^\star\).
\begin{table}[h]
\centering
\resizebox{\columnwidth}{!}{%
\begin{tabular}{lccccc}
\toprule
\textbf{Method} & \textbf{T1} & \textbf{T2} & \textbf{T3} & \textbf{T4} & \textbf{T5} \\
\midrule
CoT           & 0.2\% & 2.1\% & 1.8\% & 2.6\% & 3.8\% \\
GIoT          & 58.3\% & 62.7\% & 50.7\% & 64.2\% & 59.5\% \\
ToT           & 47.1\% & 51.4\% & 40.4\% & 63.2\% & 59.0\% \\
\rowcolor{teal!20}
TRACE (Ours)  & \textbf{84.6\%} & \textbf{87.3\%} & \textbf{82.9\%} & \textbf{93.1\%} & \textbf{90.4\%} \\
\bottomrule
\end{tabular}%
}
\caption{Coverage Ratio (averaged across 10 runs) of Generated Trajectories Across Tasks.}
\end{table}
TRACE significantly outperforms all baseline methods, achieving 83-93\% coverage across scenarios compared to 57-64\% for the next-best GIoT approach. This substantial improvement stems from TRACE's ability to systematically explore the hypothesis space through counterfactual reasoning. 
The structured nature of road environments, used in tasks T4 and T5, leads to slightly higher coverage rates for all methods compared to maritime scenarios, where the less-constrained environment admits a wider range of physically valid maneuvers. This environmental effect reveals an important property of VLM-based inference: trajectory prediction becomes more challenging precisely when the action space expands, regardless of the specific reasoning approach employed.

The performance gap between TRACE and other methods widens significantly in complex scenarios with multiple decision points. For instance, in the maritime crossing scenario (T3), TRACE outperforms GIoT by 32.2 percentage points, whereas in simpler head-on encounters (T1), this advantage narrows to 20.3 points. This pattern indicates that TRACE's counterfactual exploration provides greatest value when the target agent faces multiple viable options—precisely the situations where prediction is most crucial for downstream planning and decision-making.

\textit{Key finding 2: } \textbf{Iterative tree-of-thought expansion reveals rare but critical maneuvers.}
Beyond simple coverage metrics, TRACE demonstrates a superior ability to identify edge-case trajectories--valid maneuvers that violate typical expectations but remain consistent with constraints and observations. Figure \ref{fig:qual-figure} illustrates this distinction qualitatively. In the maritime overtaking scenario (Figure \ref{fig:qual-figure} top), where TRACE identifies unusual but legal passing trajectories including momentary accelerations before yielding and S-curve maneuvers that maintain safe distance while optimizing for faster transit time. Baselines predominantly predict canonical patterns while missing these creative alternatives that might be employed.
The road right-turn decision point (Figure \ref{fig:qual-figure} bottom) provides an even more striking demonstration of TRACE's edge case detection capability. In this scenario, the target vehicle has positioned itself with a slight rightward orientation that suggests an imminent turn. All baseline methods overwhelmingly predict right-turn trajectories, with CoT exclusively predicting a standard right turn and GIoT primarily suggesting variations of turn angle and speed. In contrast, TRACE correctly identifies that despite the suggestive positioning, the target could legitimately continue straight through the intersection by making a minor heading adjustment—a possibility confirmed in ground truth testing but missed by other approaches. This insight is valuable for safety-critical planning, as it reveals that the seemingly obvious turn intention cannot be relied upon until further committed movement is observed.
These subtleties in trajectory behavior--the ``long tail" of possible agent actions--represent precisely the challenging edge cases that most prediction systems struggle to capture but that human operators instinctively prepare for during navigation.

\begin{figure}[h]
    \centering
    \includegraphics[width=0.4\textwidth, trim=30 30 26 26, clip]{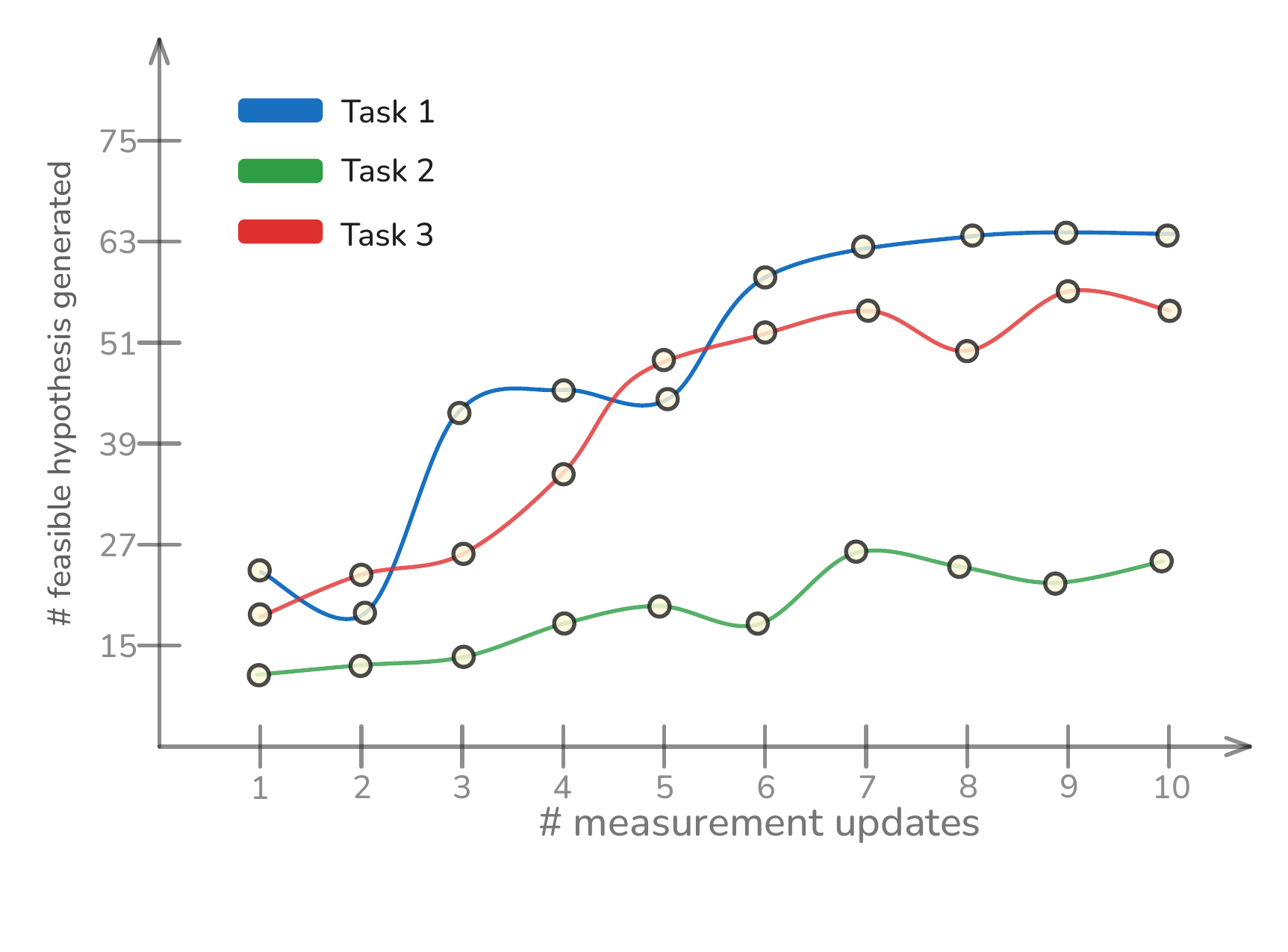}
    \caption{
Increase in VLM-generated trajectory diversity across consecutive measurement updates (averaged across 10 runs).
}
    \label{fig:graph}
\end{figure}

\textit{Key finding 3: }\textbf{VLM's outputs self-improve through iterative counterfactual exposure.}
A particularly intriguing aspect of TRACE emerges when analyzing the VLM component's output across successive measurement updates. Fig. \ref{fig:graph} shows the number of valid trajectories generated solely by the VLM's Tree-of-Thought component before any counterfactual expansions are applied. 

We observe a general upward trend in trajectory diversity as measurement updates accumulate. This trend demonstrates a powerful emergent property: the VLM appears to ``learn" from the counterfactual trajectories fed back during previous cycles. 
By the fifth measurement update, the VLM's autonomous generation capacity increases by 31.8\% across all three maritime scenarios without any explicit fine-tuning or parameter updates. 
This increase reveals that TRACE's integration mechanism creates an effective self-improving loop. When the counterfactual module identifies valid edge cases missed by the VLM, these alternatives become part of the context for subsequent VLM reasoning. The VLM then internalizes these patterns and begins proposing similar variations independently. This emergent adaptation is valuable in operational settings, where the specific behavioral patterns of a target agent become gradually more apparent over time.

\textit{Key finding 4: }\textbf{Iterative world model feedback teaches VLMs to reduce invalid trajectories.}
Beyond generating more valid trajectories, TRACE demonstrates a decrease in generation of invalid trajectories across measurement updates. Table \ref{tab-2} quantifies this effect by showing the percentage of generated trajectories that were rejected by the world model due to constraint violations or physical infeasibility.
This consistent reduction in invalid predictions occurs through the same feedback mechanism that increases valid trajectory diversity. When the world model rejects a proposed trajectory, this negative feedback is incorporated into the next round's context. The VLM gradually internalizes the domain's physical and regulatory boundaries, learning to avoid proposing maneuvers that violate kinematic constraints or environmental rules.
\begin{table}[h]
\centering
\resizebox{\columnwidth}{!}{%
\begin{tabular}{lccccc}
\toprule
\textbf{Update} & \textbf{T1} & \textbf{T2} & \textbf{T3} & \textbf{T4} & \textbf{T5} \\
\midrule
M1 & 24.8\% & 26.3\% & 25.7\% & 22.4\% & 23.9\% \\
M2 & 19.5\% & 21.7\% & 22.3\% & 18.1\% & 19.2\% \\
M3 & 14.2\% & 15.6\% & 17.8\% & 13.2\% & 14.7\% \\
M4 & 9.8\%  & 11.3\% & 12.6\% & 8.7\%  & 10.1\% \\
M5 & 7.2\%  & 8.5\%  & 9.4\%  & 6.8\%  & 7.9\% \\
\bottomrule
\end{tabular}%
}
\caption{Percentage of invalid trajectories (averaged across 10 runs) generated by TRACE across measurement updates.}
\label{tab-2}
\end{table}
The improvement is most pronounced in maritime scenarios (T1-T3), reflecting the greater complexity of maritime navigation rules and the more varied action space available in less structured environments. By contrast, road scenarios (T4-T5) begin with slightly lower invalid prediction rates and on an average achieve lower final rates, indicating that the more explicit constraints of road environments are easier for the model to internalize. 
This continual refinement is particularly valuable in operational settings, where computational resources are limited and evaluating invalid trajectories represents wasted processing. The reduction in invalid predictions directly translates to increased efficiency, allowing TRACE to allocate more computational budget toward exploring the valid trajectory space rather than filtering out impossible maneuvers.

\section{Conclusion}

This paper introduces TRACE, a novel framework that enhances Vision-Language Models for behavior hypothesis generation under sparse observations. By integrating tree-of-thought reasoning with  counterfactual exploration, we have demonstrated that VLMs can transcend their traditional limitations to produce increasingly comprehensive and robust behavior hypotheses through a self-improving cycle.

Our experimental validation across ground-vehicle simulations and real-world marine unmanned surface vehicles establishes several key findings. First, TRACE significantly improves the coverage of possible agent behaviors compared to standard VLM inference. Second, the counterfactual critic enables superior edge case detection, identifying rare but feasible maneuvers that conventional methods consistently overlook. Third, the self-improvement mechanism allows the VLM to continually enhance its output quality through iterative refinement without requiring explicit model updates. The self-improvement mechanism of TRACE also achieves a substantial reduction in invalid behavior predictions.

The implications of this work extend beyond immediate performance improvements. TRACE represents a shift in how we conceptualize the integration of foundation models into robotics—not as static prediction engines, but as reasoning components within dynamic, self-correcting systems. By demonstrating that VLMs can effectively incorporate feedback from counterfactual analysis to enhance their subsequent outputs, we establish a direction for addressing foundation models' well-documented limitations in safety-critical domains requiring comprehensive behavior representation.


\section*{Acknowledgements}
This work is supported in part by NSF grants IIS-2024733, IIS-2331908, the Office of Naval Research grant N00014-23-1-2789, N00014-23-1-2651, N00014-23-1-2505, the U.S. Department of Homeland Security grant 23STSLA00016-01-00, the U.S. Department of Defense (DoD) grant 78170-RT-REP, and the FDEP grant INV31.

\bibliographystyle{IEEEtran}
\bibliography{references}

\end{document}